\documentclass[conference]{IEEEtran}

\usepackage{algpseudocode}
\usepackage{algorithm}
\usepackage{graphicx}
\usepackage{bbold}
\usepackage{url}
\usepackage{mathtools}

\begin{document}

\title{Toward Automated Quantum Variational \\ Machine Learning}

\author{\IEEEauthorblockN{Omer Subasi}
\IEEEauthorblockA{
\textit{Pacific Northwest National Laboratory}
} Richland, Washington, USA\\
omer.subasi@pnnl.gov}

\maketitle

\pagestyle{plain}

\begin{abstract}
In this work, we address the problem of automating 
quantum variational machine learning. We develop a multi-locality parallelizable search algorithm, called MUSE,
to find the initial points and the sets of parameters that achieve the best performance for quantum variational circuit learning. 
Simulations with five real-world classification datasets indicate that on average, MUSE improves the detection accuracy of quantum variational classifiers 2.3 times with respect to the observed lowest scores.
Moreover, when applied to two real-world regression datasets, MUSE improves the quality of the predictions from negative coefficients of determination to positive ones. Furthermore, the classification and regression scores of the quantum variational models trained with MUSE are on par with the classical counterparts.

\end{abstract}


\section{Introduction}
Quantum computing \cite{rieffel2011quantum} has become the next frontier to unlock computing and information processing capabilities \cite{harrow2017quantum} that classical computing does not possess. 
These superior capabilities stem from quantum superposition and entanglement. 
As quantum-mechanical computing hardware improves, the journey to achieve its superiority over classical computing will become easier.

Meanwhile, in modern classical computing, machine learning has seen significant success throughout the last decade across many domains such as computer vision, natural language processing, robotics, autonomous decision making, generative tasks, and medical sciences \cite{dong2021survey}. Naturally, quantum machine learning has recently gained considerable attention \cite{zhang2020recent, cerezo2022challenges}. Quantum variational learning \cite{cerezo2021variational} is a main approach to quantum machine learning since it suits the current state of quantum hardware and devices, where the number of qubits is limited and quantum noise remains a significant hurdle \cite{gyongyosi2019survey}. 

In quantum variational learning, parameterised quantum circuits are utilized as models to be trained and classical computing is utilized as an optimizer to compute the loss, the gradients, and the updated values of the circuit parameters. For this reason, quantum variational learning is a hybrid quantum-classical approach. As the very first step, classical data is encoded as a unitary quantum circuit. This encoded circuit and the unitary parameterised model circuit (often called \emph{ansatz}) are then executed and the global observable representing the circuit output is measured. This output is then fed to the classical optimizer to compute the loss and the updated values of the parameters. These values are used to reconfigure the parameterised circuit and the next iteration is started by reloading (encoding) the classical data. 

In this work, we study the task of automating quantum variational learning. Existing studies related to automating quantum variational learning are mostly based on architecture (circuit) search \cite{du2022quantum, du2020quantum, NEURIPS2021_97244127, gomez2022towards, QuantumNAS, kuo2021quantum, zhang2022evolutionary, EvolutionaryQuantum, duong2022quantum}. In comparison, we propose a different approach by searching for the best combinations of initial points, circuit parameters, and data preprocessing options. Specifically, we develop a \underline{mu}lti-locality \underline{se}arch algorithm, named MUSE, to find i) the best-performing initial points, ii) the parameters of variational quantum circuits, and iii) the classical data preprocessing options. MUSE recursively searches the space of the initial points with a given combination of circuit parameters and preproccesing options, and maintains the best-performing values. As it searches the space of initial points, MUSE moves to different localities, i.e., neighborhoods. Moreover, within a locality, MUSE conducts multiple random trials by selecting different initial values in that locality. As a result, MUSE performs structured searches that significantly improve the performance of quantum variational learning as corroborated by our simulations.
In summary,
\begin{itemize}
    \item We design and develop a parallelizable search algorithm, MUSE, to automatize quantum variational learning.
    \item MUSE explores multiple localities of the space of the initial points in a methodical way based on a given set of circuit parameters and data preprocessing options to boost the performance of quantum variational learning.
    \item Simulations demonstrate that MUSE improves the detection accuracy of quantum variational classifiers 2.3 times on average with respect to the observed lowest accuracies. Moreover, MUSE improves the quality of the regression predictions from negative coefficients of determination to positive ones.
    \item Simulations also show that MUSE achieves accuracy and coefficients of determination scores that are on par with classical machine learning models.
\end{itemize}

Our study is organized as follows: 
Section \ref{background} provides the background on quantum computing and variational learning.
Section \ref{AutomatedVariational} discusses our approach to automated variational learning and our search algorithm MUSE.
Section \ref{Results} presents the simulations and results.
Section \ref{RelatedWork} surveys the related work.
Finally, Section \ref{Conclusion} concludes our work.

\begin{figure}
    \centering
    \includegraphics[width=\linewidth]{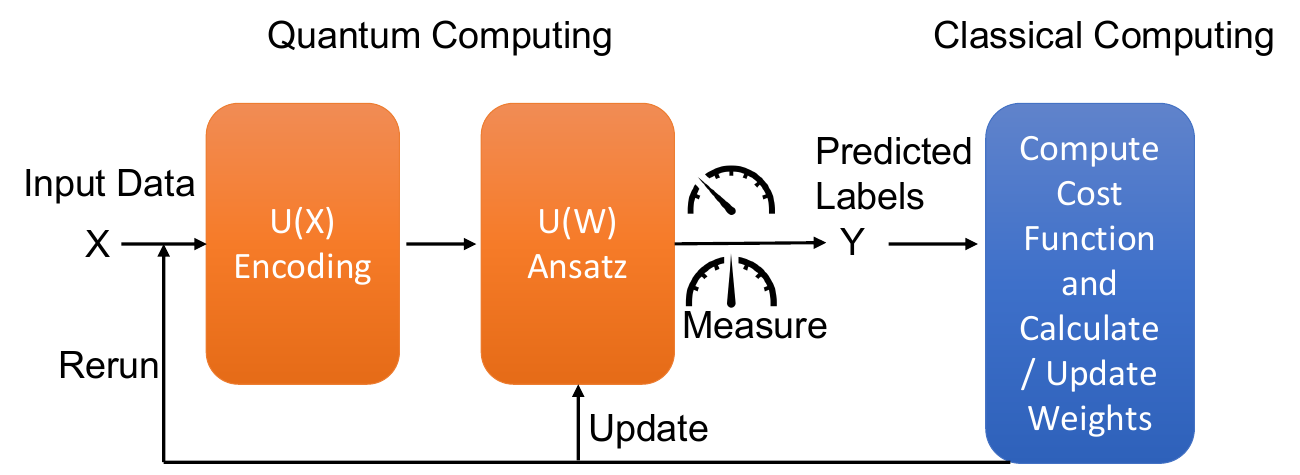}
    \caption{Quantum variational learning. $U(X)$ represents the unitary circuit to encode the classical input $X$. $U(W)$ represents the unitary quantum circuit parameterized by the trainable weights $W$ which is often called ansatz.}
    \label{fig:variationallearning}
\end{figure}

\section{Background}
\label{background}
\subsection{Quantum Computing: Basic Formalism}
The (pure) state of qubit is formalized by $|\psi \rangle = \alpha |0 \rangle + \beta |1 \rangle$ where amplitudes $\alpha$ and $\beta$ are complex numbers, and 
$|\alpha|^2 + |\beta|^2 = 1$. In particular, the computational basis is $|0 \rangle = (1, 0)^T$ and $|1 \rangle = (0, 1)^T$ where the superscript $T$ denotes the transpose of a vector.
Multiple $N$-qubit systems are formed by the tensor product of
$N$ single qubit systems. 
A single-qubit operation, called a \emph{gate} in the circuit model of quantum computing, is represented by a $2 \times 2$ matrix, such as the Pauli matrices
$ 
X = \left( \begin{array}{cc} 0 & 1 \\
1 & 0 \end{array} \right), 
Y = \left( \begin{array}{cc} 0 & -i \\
i & 0 \end{array}\right), $ and
$
Z = \left( \begin{array}{cc} 1 & 0 \\
0 & -1 \end{array} \right)
$.
Parameterized rotation gates $R_X(\theta) = e^{- i \theta X}$, 
$R_Y(\theta) = e^{- i \theta Y}$ and $R_Z(\theta) = e^{- i \theta Z}$, and phase gates $P(\theta)$ are other common single-qubit operations. 
Moreover, 
Hadamard gate $H = \frac{1}{\sqrt{2}}  \left( \begin{array}{cc} 1 & 1 \\
1 & -1 \end{array} \right)$ places a qubit in perfect superposition. 
The two-bit controlled-not gate
$ \\
CZ = \left( 
\begin{array}{cccc} 1 & 0 & 0 & 0\\
0 & 1 & 0 & 0 \\ 
0 & 0 & 1 & 0 \\ 
0 & 0 & 0 & -1 
\end{array} 
\right)
$
is used to entangle two qubits.\\

Measurements are performed to obtain information from a quantum system. Due to Born's rule, measuring a qubit state $|\psi \rangle = \alpha |0 \rangle + \beta |1 \rangle$ produces $0$ with probability $|\alpha|^2$ and  
$1$ with probability $|\beta|^2$. A measurement operation, also called an \emph{observable}, is mathematically represented by a Hermitian matrix whose eigenvalues are real - a necessary condition for real physical experiments. An observable can be either \emph{local} or \emph{global}. If the observable acts on all qubits, it is global. Otherwise, it is local.
Due to probabilistic nature of quantum mechanics, to get statistically significant results, multiple measurements need to be repeated to calculate the average (expectation) of an observable $O$
\begin{equation} 
    \langle \psi | O | \psi  \rangle  \equiv Tr[O |\psi \rangle \langle \psi|] \nonumber
\end{equation}
where $Tr$ denotes the trace of a matrix.

\subsection{Quantum Variational Machine Learning}
In this section, we overview the parameterized (variational) quantum circuits. 
Parameterized circuits allow quantum machine learning algorithms to be developed and tested on the near-term devices \cite{cerezo2021variational, gyongyosi2019survey}.
In parameterized circuits, the quantum gates are defined through trainable parameters and training aims to optimize these parameters.
Training of these circuits follows a number of steps that are based on both classical and quantum computing in a variational manner. As a result, the training process is called quantum variational learning and is a hybrid approach.
As the very first step, classical input data is encoded into a quantum circuit which is called the \emph{feature map circuit}. Data encoding can be done in different ways such as basis, amplitude, angle, and arbitrary encoding \cite{schuld2021machine, schuld2018supervised}. Once the classical input data is encoded into a quantum circuit, the parameterized circuit with trainable weights is executed.
This circuit is often called \emph{ansatz}, which is typically implemented by single-qubit rotation gates. After this step, the resulting quantum output state is measured. The measurement produces classical data which is then fed to a classical optimizer. The classical optimizer defines a cost function and computes the loss. It then performs the optimization and provides the updated parameters of the circuit. Finally, the initial classical data is reloaded (encoded) and the circuit is executed in the next iteration. Figure \ref{fig:variationallearning} illustrates the steps of quantum variational learning. 

Formally, assuming that for the classical input $X$, $U(X; \Theta_E)$ with parameters $\Theta_E$ and $U(X; W(\Theta_A))$ with weight parameters $\Theta_A$ represent the unitaries for the encoding (feature map) and the ansatz respectively, we build a quantum variational circuit $\Psi$ from the ground (zero) state
\begin{equation} \nonumber
   |\Psi(X; \Theta_E \cup \Theta_A) \rangle =  U(X; W(\Theta_A)) U(X; \Theta_E) |0 \rangle.
\end{equation}
The training task therefore becomes minimizing the average 
\begin{equation} \nonumber
   \langle |\Psi(X; \Theta_E \cup \Theta_A) |O|\Psi(X; \Theta_E \cup \Theta_A)  \rangle
\end{equation}
for a global observable $O$, which represents the values of the circuit's output qubits. In quantum machine learning, observables are typically simple. For instance, for classification tasks, the observable can be $|0 \rangle \langle 0|$ for a single output qubit $0$. 

\begin{figure*}[!ht]
    \centering
    \includegraphics[width=\linewidth]{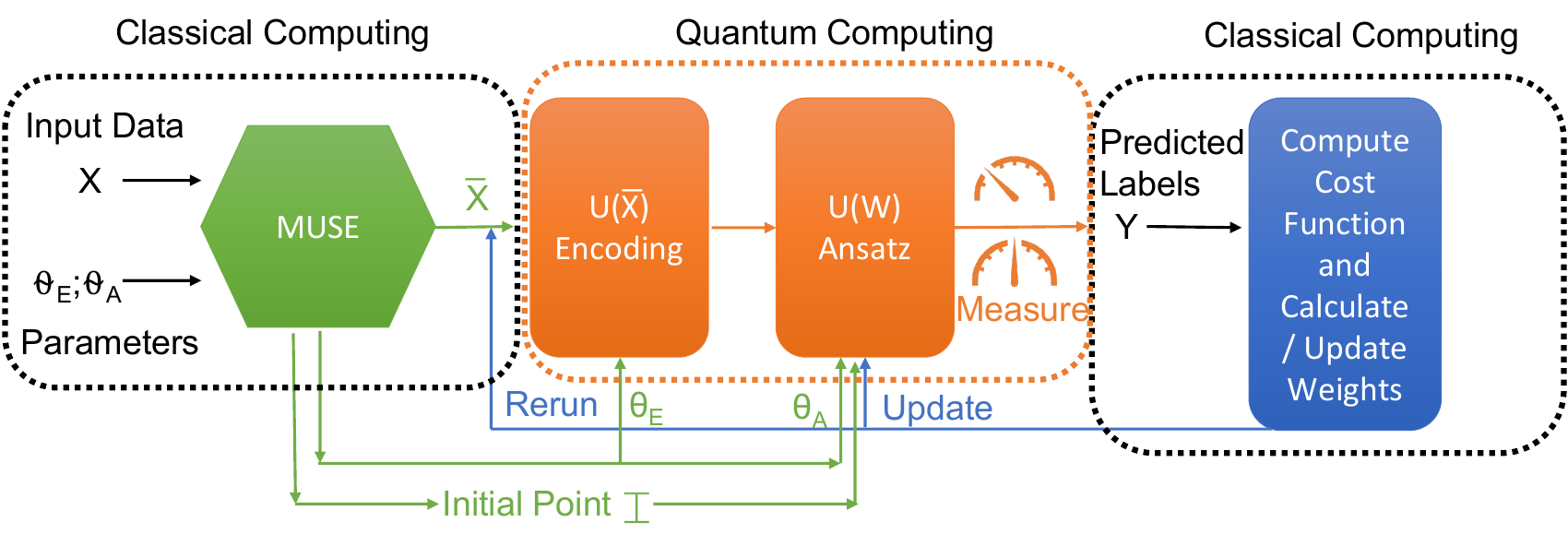}
    \caption{MUSE as a component of hybrid quantum-classical variational learning. $\overline{X}$ is the transformed input ${X}$ based on the classical data preprocessing options that MUSE evaluates.
    $U(\overline{X})$ represents the unitary encoding circuit (feature map) for the input $\overline{X}$. $U(W)$ represents the unitary ansatz circuit having the trainable weights $W(\Theta_A)$. 
    $\Theta_E$ and $\Theta_A$ correspond to the collection of the input parameters of the encoding and the ansatz circuit respectively.  MUSE selects the initial point I and the parameters $\theta_E$ and $\theta_A$ as inputs to the encoding circuit $U(\overline{X})$ and the ansatz circuit $U(W)$.}
    \label{fig:muse}
\end{figure*}

Successful machine learning models generalize well. In quantum variational learning, parameterized variational quantum circuits are utilized as the models. To characterize and quantify how well quantum variational circuits generalize, the concepts of \emph{expressibility} and \emph{entangling capability} have been proposed by Sim, Johnson and Aspuru-Guzik \cite{sim2019expressibility}. The expressibility of a variational circuit refers to the extent to which the circuit can generate different states within the Hilbert space. The authors \cite{sim2019expressibility} quantify expressibility by computing the difference between the distribution of the states generated from the circuit and the uniform distribution. Expressive circuits represent many different unitaries. 
On the other hand, the entangling capability of a variational circuit describes its ability to generate entangled states. The authors \cite{sim2019expressibility} quantify entangling capability by Meyer-Wallach measure. Highly entangled quantum circuits behave close to the Bell state.

\begin{algorithm}[!ht]
\caption{MUSE Search Algorithm}\label{alg:searchalg}
\begin{algorithmic}[1]
\Procedure{MUSE }{$best\_pt,  best\_sc,  args$} 
\State $\epsilon, \alpha, \beta, \mathbb{U}, \mathbb{L}, depth, params \gets args$ \label{l1a}
\State $prev \gets best\_sc$
\If{$(depth \leq 0)$}
    \State \textbf{return} $best\_pt, \  best\_sc, \ args$
\Else 
 \State $depth \gets depth - 1$
\EndIf \label{l1b}
\State $\epsilon_l \gets$ \textbf{random$(low=0, high=\epsilon)$} \label{l2a}
\State $\epsilon_r \gets$ \textbf{random$(low=0, high=\epsilon)$}
\State  $lpt \gets $ \textbf{min}((\textbf{max}($\mathbb{U} - best\_pt, \mathbb{L}$) + $\epsilon_l$), $\mathbb{U}$)
\State  $rpt \gets \textbf{min}(\mathbb{U}, best\_pt + \epsilon_r)$
\State $lscore, lpt \gets \textbf{run} (lpt, params)$ \label{l2b-init}
 \State $rscore, rpt \gets \textbf{run} (rpt, params)$ \label{l2b}
\If{$(lscore > best\_sc)$} \label{l3a}
 \State       $best\_pt \gets lpt$ 
 \State       $best\_sc \gets lscore$
 \EndIf
\If{$(rscore > best\_sc)$}
 \State       $best\_pt \gets rpt $
  \State      $best\_sc \gets rscore$
\EndIf \label{l3b}
\If{$(prev < best\_sc)$} \label{l32a}
\State $args \gets \{\epsilon, \alpha, \beta, \mathbb{U}, \mathbb{L}, depth, params\}$
\State \textbf{return} $\textbf{MUSE}\ (best\_pt, best\_sc, args)$
\EndIf \label{l32b}
\If{$(prev == best\_sc)$} \label{l4a}
\State   $tmp\_pt \gets \alpha \times best\_pt$
\State $ score, pt \gets \textbf{run}(tmp\_pt, params)$
\If{$(score > best\_sc)$}
\State $best\_pt = pt$ 
\State $best\_sc = score$
\State $ depth \gets depth-1$
\State $args \gets \{\epsilon, \alpha, \beta, \mathbb{U}, \mathbb{L},  depth, params\}$
\State \textbf{return} $\textbf{MUSE}\ ( best\_pt, best\_sc, args)$
\EndIf
\EndIf \label{l4b}
\If{$(prev > best\_sc)$}
\State   $tmp\_pt \gets \beta \times best\_pt$
\State $ score, pt \gets \textbf{run}(tmp\_pt, params)$
\If{$(score > best\_sc)$} \label{l5a}
\State $best\_pt = pt$ 
\State $best\_sc = score$
\State $ depth \gets depth-1$
\State $args \gets \{\epsilon, \alpha, \beta, \mathbb{U}, \mathbb{L}, depth, params\}$
\State \textbf{return} $\textbf{MUSE}\ ( best\_pt, best\_sc, args)$
\EndIf
\EndIf \label{l5b}
\State \textbf{return} $best\_pt, \  best\_sc, \ args$
\EndProcedure
\end{algorithmic}
\end{algorithm}

\section{MUSE for Automated Quantum Variational Learning}
\label{AutomatedVariational}
In this section, we elaborate on our approach to automatizing
quantum variational learning and the MUSE algorithm.

\subsubsection{Our Approach}
Our automation approach is based on two parts. 
The first part is our multi-locality recursive search algorithm MUSE.
In MUSE, we search for an initial point to train a quantum variational and parameterised ansatz circuit such that the performance on the test data is the highest. We search for such point starting with a random point uniformly chosen from the value space that is fully described by an upper and lower bound. We then methodically explore this space. 
Given an initial random point, we explore two different points: The first point is close to this initially selected point. That is, a point within the vicinity of the first point. In MUSE, we use a hyper-parameter $\epsilon$ to quantify the vicinity, i.e, the neighborhood, to a point.
The second point is selected by a point closer to the ``opposite side'' of the value space - closer to the upper or lower bound depending on the value of the preceding initial point. Practically,
because in classical machine learning it is very common to normalize or standardize the data, each data feature is between the lower bound of 0 and the upper bound of 1. Therefore, typically, [0, 1] is the interval for a single feature dimension. When the data is multi-dimensional, we treat each dimension the same. 

After determining the two points, we observe the performance score of the variational circuit for both of them. If there is an improvement in performance, we set the best-performing point as the next initial point and we continue the recursive search. We also save this best-performing point as we continue searching.
The length of the search is controlled by a \emph{depth} parameter which controls the number of recursive calls to MUSE.
If there is no performance improvement, however, one of the two points performs the same as the best-performing point so far, then we make another attempt with scaling the best-performing point. Controlled with a scale hyper-parameter $\alpha$, we evaluate the variational circuit at the scaled point. If there is an improvement in performance, then we move to the next recursive call if the depth of the search permits. If not, then we return the best-performing point.
In the case where both points under perform, we scale the the best-performing point by a hyper-parameter $\beta$ ($< \alpha$ due to under-performance) and feed the scaled point to the quantum circuit to observe the performance score. If the score increases and the depth of the search permits, we continue the search. Otherwise, we return the best-performing point. We note that we conservatively devise MUSE to immediately terminate a recursive call if there is no performance improvement after the third trial. This is to facilitate searching in as many localities as possible provided that the depth limit permits. 

The second part to our automation approach is to explore and evaluate a predetermined set of parameter combinations and options with MUSE. These can be the parameters that define the variational circuit or the options that determine the type of classical data preprocessing, such as MinMax or Standard normalization, and the method for dimensionality reduction, such as principal component analysis or univariate feature selection. For each parameter and option combination, we search for a best-performing initial point. In the end, we find the best-performing parameter values, the preprocessing options, and the initial point to the variational circuit. Figure \ref{fig:muse} illustrates MUSE as part of hybrid quantum-classical variational learning. 

\subsubsection{MUSE Pseudocode}
Algorithm \ref{alg:searchalg} presents the pseudocode of MUSE. Lines \ref{l1a} - \ref{l1b} show that MUSE stores the best-performing point and checks for termination based on the depth parameter.
Lines \ref{l2a} - \ref{l2b} run the quantum circuit to get the scores for the given initial point and parameters.
Lines \ref{l3a} - \ref{l3b} check if there is an improvement in the scores as MUSE maintains the best-performing point and parameters.
Lines \ref{l32a} - \ref{l32b} make a subsequent MUSE call if there is a performance improvement.
Lines \ref{l4a} - \ref{l4b} check if the current initial point performs the same as the best-performing initial point and executes the circuit again. If there is some improvement in performance, a subsequent MUSE call is initiated. If not, the current call is ended with the best-performing initial point and parameters.
Lines \ref{l5a} - \ref{l5b} take the same steps when there is no performance improvement.

We can see from the lines \ref{l2b-init} - \ref{l2b} that certain sections of MUSE is easily parallelizable, which helps reduce the wall-time of MUSE instantiations. 

\begin{figure}[ht]
    \centering
    \includegraphics[width=\linewidth]{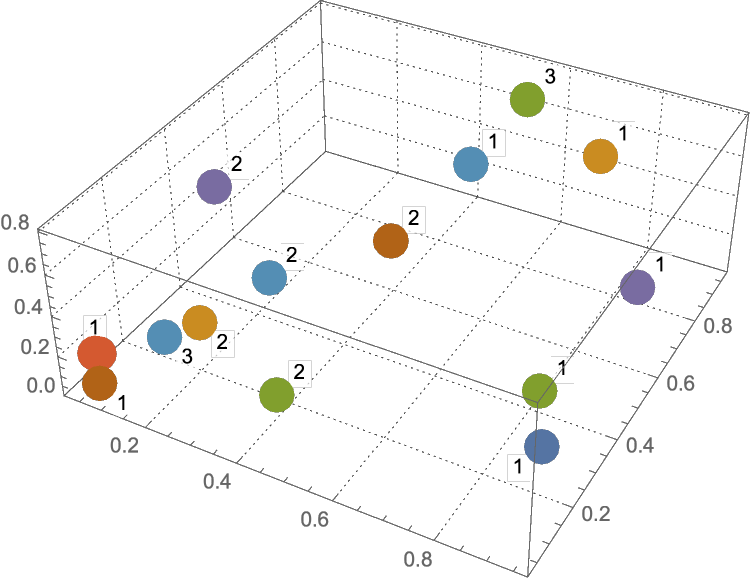}
    \caption{An example instantiation of MUSE. We see that MUSE can explore one to three localities in a single run - that is, not always three. In some cases, it only explores two localities by moving between them. In one case, colored with red on the bottom left corner, MUSE explores only one locality as it runs out of the permitted depth of three. This means that MUSE randomly started from a locality that has high-performing initial points by chance.
    }
    \label{fig:3dscatter}
\end{figure}

\subsubsection{Illustration}
We design MUSE so as to search three localities given an initial point $X \in \mathbb{R}^n$ and the locality (neighborhood/vicinity) determined by the parameter $\epsilon \in 
 \mathbb{R}$. Formally, these are the union of the hyperspheres $S_n (X, \epsilon)$ and $S_n (\alpha X, \epsilon)$, the hypersphere $S_n (U-X, \epsilon)$, and the hypersphere $S_n (\beta X, \epsilon)$. Here, $S_n (X, \epsilon)$ denotes the hypersphere ($n$-hypersphere) centered at the $n$-dimensional point $X$ with a radius of $\epsilon$. To illustrate these localities, Figure \ref{fig:3dscatter} visualizes an example instantiation of MUSE. In this run, we assume a three dimensional input to a quantum variational circuit where the input is normalized to [0, 1] and the permitted depth is three. In the figure, we use different colors to show different instantiations of MUSE. The numbers on the same colored points show the order of exploration by MUSE. We see that MUSE can explore up to three localities in a single instantiation - that is, not always three. In some cases, it only explores two localities by moving between them. In one case, colored with red on the bottom left corner, MUSE explores only one locality as it runs out of the permitted depth of three. This means that MUSE randomly started from a locality that has high-performing initial points by chance.

\section{Simulations and Results}
\label{Results}
In this section, we detail our simulations and discuss the results. In particular,
we first present the experimental setup and the datasets.
We then analyze the results of the simulations.
After that, we present two of our earlier attempts that failed to improve the
performance of the variational learners. 

\subsection{Experimental Setup and Datasets}
We implement MUSE and the auxiliary codes in Qiskit \cite{Qiskit} and Qiskit Machine Learning. Table \ref{tab:softwareversions} shows the relevant software versions.
The simulations were run at the Deception Computing System at PNNL using single compute nodes. 
A compute node has
dual AMD EPYC 7502 CPUs running at 2.5 GHz with the ability to boost to 3.35 GHz, 
256 GB of octa-channel DDR4-3200 memory,
512 GB of local NVMe.

\begin{table}[ht]
    \centering
    \caption{Software Details}
    \label{tab:softwareversions}
        \begin{tabular}{|l|l|}  
     \hline
\textbf{Software} &	\textbf{Version} \\ \hline
qiskit-terra &	0.23.2 \\ \hline
qiskit-aer	& 0.12.0 \\ \hline 
qiskit-ibmq-provider	& 0.20.2 \\ \hline
qiskit	& 0.42.0 \\ \hline
qiskit-machine-learning	& 0.5.0 \\ \hline
Python version &	3.9.5 \\ \hline
    \end{tabular}    
\end{table}
\begin{table*}[!ht]
    \centering
    \caption{Classification Datasets}
    \label{tab:datasets}
        \begin{tabular}{|l|l|l|l|l|l|}  
     \hline
     Datasets & \textbf{Iris} & \textbf{Cancer} & \textbf{Wine} & \textbf{Diabetes} & \textbf{Blood} \\ \hline
Classes & 3 & 2 & 3  & 2 &  2 \\ \hline
Samples per class & [50, 50, 50] & [212, 357] & [59, 71, 48] & [500, 268] & [570, 178] \\ \hline
Samples total & 150 & 569 & 178 & 768 &  748\\ \hline
Dimensionality & 4 & 30 & 13 & 8 & 4 \\ \hline
Features & Real, positive & Real, positive & Real, positive & Real, positive  & Real, positive \\ \hline
    \end{tabular}    
\end{table*}
\begin{table}[!ht]
    \centering
    \caption{Regression Datasets}
    \label{tab:Regressiondatasets}
        \begin{tabular}{|l|l|l|}  
     \hline
     Datasets & \textbf{Diabetes (Regression)} & \textbf{Liver Disorders}   \\ \hline
Samples total  & 442 & 345 \\ \hline
Dimensionality  & 10 & 5 \\ \hline
Features  & Real & Integer, positive\\ \hline
Target  & Integer, positive  & Real, non-negative \\ \hline
    \end{tabular}    
\end{table}
\begin{table}[!ht]
    \centering
    \caption{Simulation Parameters \\ (Fixed and not comparatively evaluated)}
    \label{tab:simparams}
        \begin{tabular}{|l|l|}  
     \hline
     \textbf{Parameter} & \textbf{Value}   \\ \hline
     Feature Map & ZZFeatureMap \\ \hline
     Ansatz & RealAmplitudes \\ \hline
Train-Test Proportion  & 80\%-20\% \\ \hline
Dimensions after feature reduction (classification)  & 4 \\ \hline
Dimensions after feature reduction (regression)  & 2 \\ \hline
Classification optimizer & COBYLA \\ \hline
Regression optimizer & L\_BFGS\_B \\ \hline
Classification iterations & 100 \\ \hline
Regression iterations & 10 \\ \hline
Number of trials  & 2 \\ \hline
$\epsilon$  & 0.02 \\ \hline
$\alpha$  & 0.9 \\ \hline
$\beta$  & 0.5 \\ \hline
Search depth & 3 \\ \hline
    \end{tabular}    
\end{table}

The classification datasets which are used in our simulations are the Iris Plant \cite{Dua2019}, the Breast Cancer Wisconsin \cite{Dua2019}, the Wine \cite{Dua2019}, the Diabetes \cite{Dua2019} and the Blood \cite{yeh2009knowledge} datasets.
Table \ref{tab:datasets} provides details about them.

The regression datasets are the Liver Disorders \cite{mcdermott2016diagnosing} dataset and the Diabetes Regression dataset \cite{Efron2004} \cite{ScikitLearnDiabetes}.
Table \ref{tab:Regressiondatasets} shows details about these two datasets.

Table \ref{tab:simparams} details the simulation parameters and their values used in our simulations. All other parameters are set with the default values. We note that these simulation parameters are not searched and not comparatively evaluated. They are fixed for all simulations. The parameters that are searched by MUSE are discussed below and are shown in Algorithm \ref{alg:driver}. In our simulations, we use preset seeds to obtain reproducible results. Each driver run is repeated twice with different seeds to see if the results are consistent. With the seeds we simulate, the results are consistent. Lastly, all simulations are performed by the built-in Qiskit simulators.

\begin{algorithm}[ht]
\caption{Driver Procedure}\label{alg:driver}
\begin{algorithmic}[1]
\Procedure{Driver}{}
\State $n\_trials \gets 2$ \label{a1}
\State $\epsilon \gets 0.02$
\State $\alpha \gets 0.9$
\State $\beta \gets 0.5$
\State $\mathbb{U} \gets \mathbb{1}$
\State $\mathbb{L} \gets \mathbb{0}$
\State $depth \gets 3$
\State $best\_sc \gets 0$
\State $best\_pt \gets \mathbb{0}$
\State $best\_params \gets [] $ \label{b1}
\State $feat\_ans \gets [(1, 2), (1, 3), (1, 4), (2, 3), (2, 4)]$ \label{a2}
\State $sca\_red \gets [ (std, pca), (std, f),(mm, pca), (mm, f) ]$\label{a3}
\For{$i=1$ \textbf{to} $n\_trials$} \label{a4}
\For {$feat, ans \in feat\_ans$}
\For {$sca, red \in sca\_red$}
\State $pt \gets \textbf{random()}$
\State $params \gets \{feat, ans, sca, red\}$
\State $args \gets \{\epsilon, \alpha, \beta, \mathbb{U}, \mathbb{L}, depth, params\}$
\State  $pt, sc, args \gets \textbf{MUSE} (pt, sc, args)$
\If{$sc > best\_sc$} \label{a5}
    \State $best\_sc \gets sc$
    \State $best\_pt \gets pt$
    \State $best\_params \gets params$
\EndIf \label{b5}
\EndFor
\EndFor
\EndFor \label{b4}
\State \textbf{print}('best init point', $best\_pt$) 
\State \textbf{print}('best score', $best\_sc$)
\State \textbf{print}('best args', $best\_params$) 
\EndProcedure
\end{algorithmic}
\end{algorithm}

Algorithm \ref{alg:driver} shows an example driver procedure. 
The parameters within the driver are searched and comparatively evaluated by MUSE.
The procedure first initializes the hyper-parameters of MUSE (Lines \ref{a1} - \ref{b1}). We assume that the (multi-dimensional) data is normalized, that is, $\mathbb{U} = \mathbb{1}$ and $\mathbb{L} = \mathbb{0}$ where $\mathbb{1}$ and $\mathbb{0}$ represent the multidimensional matrices with all elements being 1 and 0 respectively.
Line \ref{a2} sets the parameter grid that determines the quantum variational circuit. A pair of numbers in this grid refers to the number of repetitions in the feature map and the ansatz of the parameterized quantum circuit to be evaluated. The numbers of repetitions are chosen such that we can simulate them in a reasonable time.
Line \ref{a3} sets the parameter grid for classical data preprocessing. Here, \textit{std} refers to standard scaling, \textit{mm} refers to MinMax scaling, 
\textit{pca} refers to principal component analysis, and \textit{f} refers to univariate feature selection based on the ANOVA F-value for feature (dimensionality) reduction. Lines \ref{a4} - \ref{b4} iteratively go through the two parameter grids, select a random initial point and launch MUSE. After every iteration, Lines \ref{a5} - \ref{b5} save the best-performing initial point, parameters and score yet.

\subsection{Simulation Results}
In this section, we present the results of our simulations for MUSE with quantum variational learners for classification and regression. We compare MUSE to classical classification and regression. We also compare MUSE to the parameter search performed with random initializations, which we name as \textit{Random} search. MUSE can run a variational circuit by up to 2$\times$ the depth parameter in a search. For Random search, we always run a circuit 2$\times$ the depth. This is unfair to MUSE, however, it provides an opportunity that reveals how competitive MUSE can be. Additionally, we note that MUSE achieves the performances that we report by just two initializations (See Table \ref{tab:simparams} and Algorithm \ref{alg:driver}). Even with only two initializations, due to the large number of the combinations of the parameters and data preprocessing options, and the repeated driver runs with different seeds, the total simulation time is about 14000 CPU hours in this study.

\subsubsection{Classification Results}
\begin{figure}[ht]
    \centering
    \includegraphics[width=\linewidth]{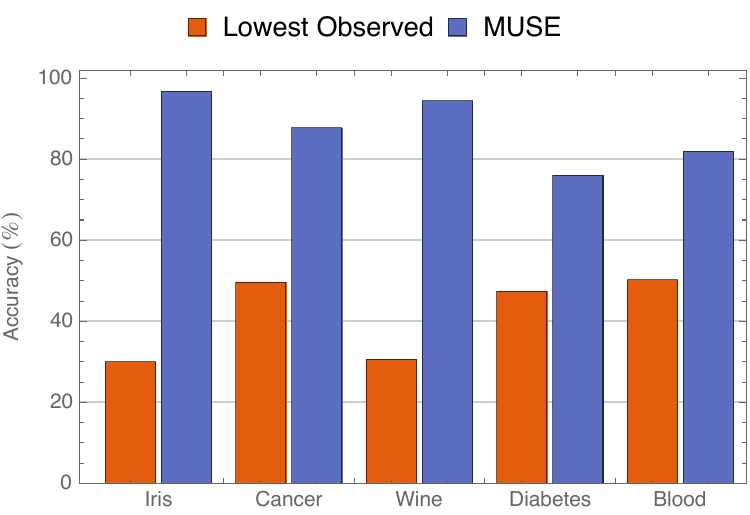}
    \caption{The accuracy scores achieved by MUSE and the lowest accuracy scores observed during the MUSE search.
    }
    \label{fig:minmaxscores}
\end{figure}

\begin{figure}[ht]
    \centering
    \includegraphics[width=\linewidth]{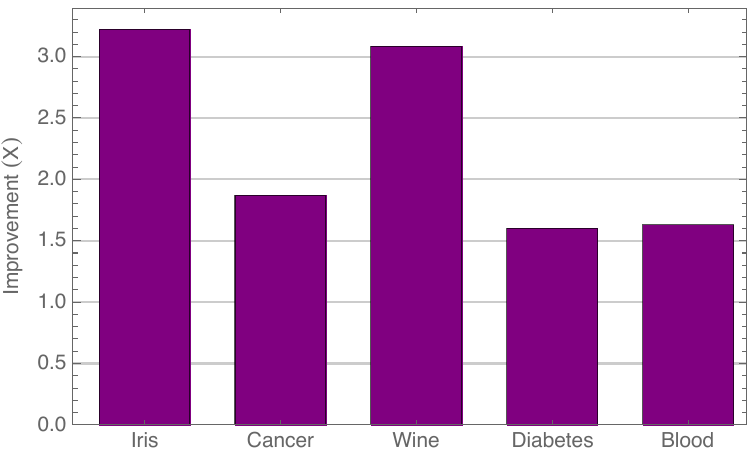}
    \caption{Accuracy improvements with MUSE in comparison to the lowest accuracy scores observed during the MUSE search.}
    \label{fig:improvements}
\end{figure}

Figure \ref{fig:minmaxscores} shows the accuracy scores achieved by MUSE and the lowest scores occurred during the search for each dataset. Without MUSE, random initializations could lead to such lowest scores. From the figure, 
we see that for all five datasets, MUSE significantly improves the detection accuracy over the lowest scores observed. The accuracy improvements are as large as 97\% with respect to the lowest score of 30\% observed in the Iris dataset. Figure \ref{fig:improvements} depicts the accuracy improvements achieved by MUSE with respect to the observed lowest accuracy scores for each dataset. We see that the accuracy improvements get as high as more than 3$\times$. On average, MUSE improves the detection performance of the quantum variational classifiers 2.3$\times$. 

Table \ref{tab:bestworstparams} shows the best- and worst-performing parameter combinations for the quantum variational classifiers. We see that the MinMax scaler is consistently the best-performing scaler. We also see that F-ANOVA based feature selection (FA) most often outperforms principal component analysis based feature reduction (PCA). In terms of the number of repetitions of the feature map and the ansatz, no particular combination outperforms the others.

\begin{table*}[ht]
    \centering
    \caption{Classification: Best and worst performing parameter combinations}
    \label{tab:bestworstparams}
    \begin{tabular}{|l|l|l|l|l|l|l|l|l|}  
     \hline
     &
     \multicolumn{4}{|c|}{\textbf{Best Parameters}} &
     \multicolumn{4}{|c|}{\textbf{Worst Parameters} } \\ \hline
        Dataset  &  ZZ Map Reps. & Ansatz Reps. & Scaler & Reducer &  
 ZZ Map Reps.  & Ansatz Reps. & Scaler & Reducer \\ \hline
       \textbf{Iris}  & 2  & 3 &  MinMax & FA & 2  & 3 &  Standard & FA  \\ \hline
       \textbf{Cancer}  &  1  & 4 &  MinMax & FA & 1  & 3 &  Standard & PCA   \\ \hline
       \textbf{Wine}  &  1  & 4 &  MinMax & FA & 2  & 3 &  Standard & PCA  \\ \hline
       \textbf{Diabetes}  &  2  & 3 &  MinMax & FA & 1  & 4 &  Standard & FA   \\ \hline
      \textbf{ Blood}  & 1  & 4 &  MinMax & PCA & 1  & 3 &  Standard & PCA \\ \hline
    \end{tabular}
    
\end{table*}

\begin{figure}
    \centering
    \includegraphics[width=\linewidth]
    {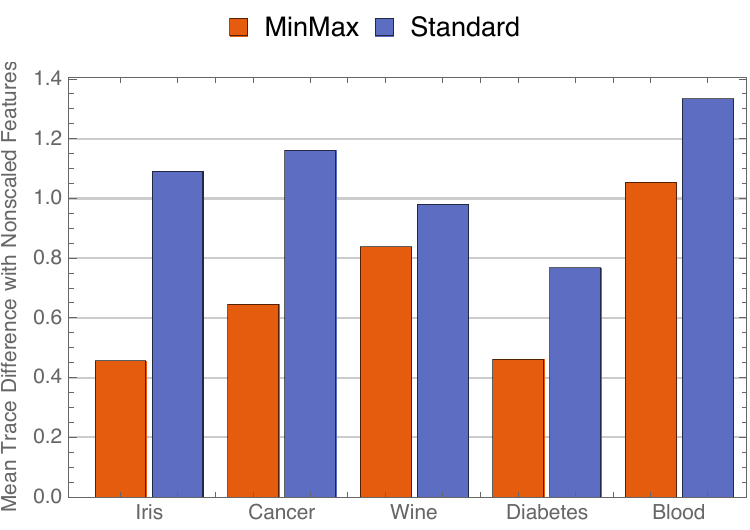}
    \caption{The mean trace differences of MinMax and Standard scaling's phase rotation matrix values with the reference matrix values.}
    \label{fig:traceminmax}
\end{figure}

\begin{figure}
    \centering
    \includegraphics[width=\linewidth]{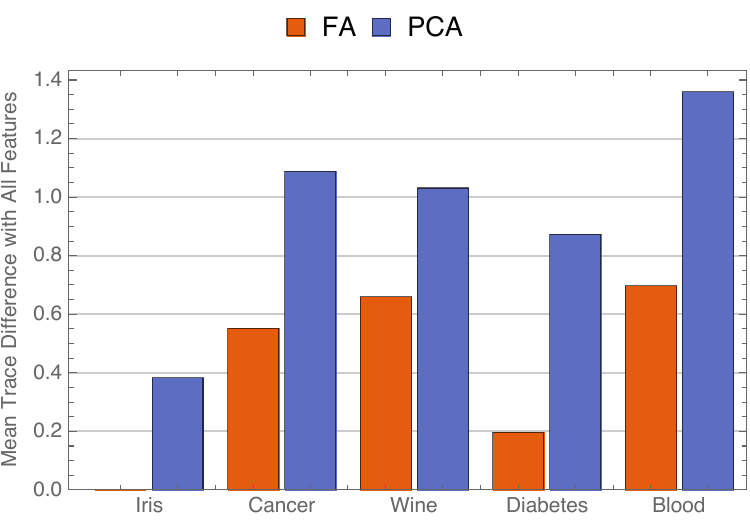}
    \caption{The mean trace differences of FA-based and PCA-based circuits' output matrix values with the reference matrix values.}
    \label{fig:tracefa}
\end{figure}

To understand why MinMax scaler and FA feature reduction performs better than Standard scaler and PCA based feature reduction, we construct separate short test quantum circuits that have inputs preprocessed by the two scalers and the two feature reducers. We also construct a reference circuit that has the original unprocessed inputs. All test circuits have the same structure: for each qubit, a Hadamard gate is followed by a phase gate with the respective (scaled or reduced) features as their input parameters. We use phase gates because the feature map i.e., the encoding circuit, we employ is the \emph{ZZFeatureMap}, and it is constructed by phase gates in Qiskit.
We run the circuits with the Qiskit unitary simulator after which we get the resulting output matrices. We then compute the mean trace differences with the reference output. Figures 
 \ref{fig:traceminmax} and \ref{fig:tracefa} show the mean trace differences with the reference values. In particular, Figure \ref{fig:traceminmax} shows that MinMax scaling produces closer phase rotation values with the reference circuit. On the other hand, Standard scaling produces values - including negative - that are relatively more distant from the reference values thereby causing lower accuracy. In terms of the impact of feature reduction type, Figure \ref{fig:tracefa} compares FA and PCA based feature reduction. We see that FA based reduction leads to the rotation values that are closer to the reference circuit. FA is a \emph{feature selection} algorithm which is completely deterministic. This means that FA does not produce new classical data features or transform them; it only selects a subset of them. In contrast, PCA is a \emph{feature extraction} method. It finds the principal and orthogonal components in the data and projects the entire data on these newly found components, which are some combination of the original features. Therefore, PCA naturally produces more distant values than the reference rotation values as it transforms the data. This, in turn, leads to lower accuracy performance. 
 
 As an experimental note, we set the target number of reduced features to 3 (not 4 as in Table \ref{tab:simparams}) for the comparison of FA and PCA - and only for this comparison. This is because when the number of features is 4, then FA simply produces to the same values with the reference values for all datasets. We lower the dimension to 3 to get nonzero difference between FA and the reference values. As seen in Figure \ref{fig:tracefa}, even with 3 dimensions, FA still produces to the same values with the reference ones for the Iris dataset.

\begin{figure}[ht]
    \centering
    \includegraphics[width=\linewidth]{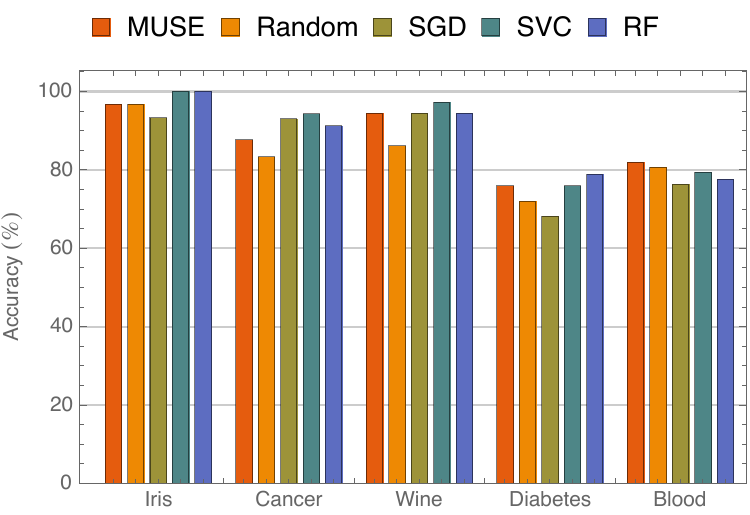}
    \caption{Comparison of the accuracy performance of the QVCs trained with MUSE, the QVCs trained with Random search, and the classical classifiers.}
    \label{fig:classres}
\end{figure}

Figure \ref{fig:classres} compares the accuracy performance of the quantum variational classifiers (QVCs) trained with MUSE to the three classical classifiers and to the QVCs trained with Random search. The three classifiers are a linear stochastic gradient descent (SGD) based classifier, a Support Vector Machines (SVMs) based classifier (SVC), and a Random Forest (RF) classifier. We clearly see that the QVCs trained with MUSE achieve accuracy performances that are on par with the classical classifiers. 

From Figure \ref{fig:classres}, we see that the QVCs trained with MUSE outperform the QVCs trained with Random search in all datasets except the Iris dataset for which they perform the same. However, the performance differences are limited. In particular, 
MUSE outperforms Random search by 4.4\%, 8.3\%, 4\%, and 1.3\% for the Cancer, Wine, Diabetes, and Blood datasets respectively.
This is partially because the expressive power of the tested circuits seems to be nearly fully utilized. For all datasets, the highest achievable performance seems to be reached. This explanation is further supported by the similar performance of the classical classifiers. Another reason is that Random search tests QVCs more times than MUSE does - as we stated earlier.

\begin{figure}
    \centering
    \includegraphics[width=\linewidth]{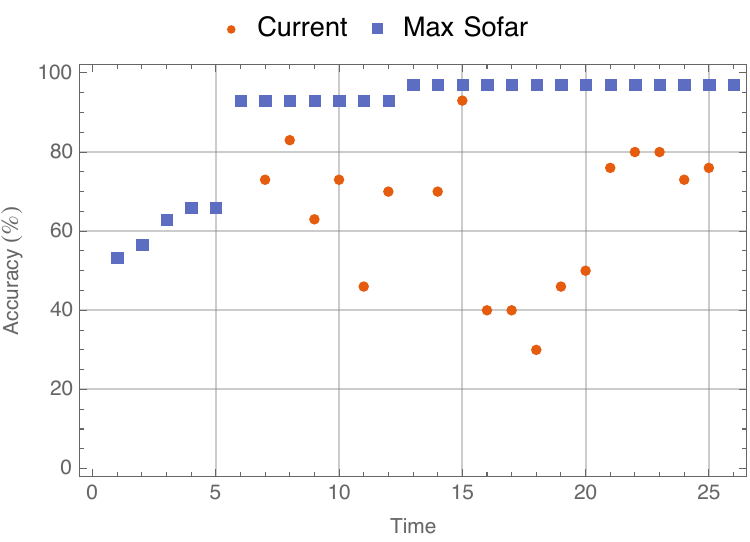}
    \caption{An instantiation of MUSE with the Iris dataset. The plot shows the maximum accuracy achieved yet and the accuracy progression throughout the instantiation. The Time axis is not in any real physical unit. The axis rather represents a conceptual time passage.}
    \label{fig:progress}
\end{figure}

Figure \ref{fig:progress} shows a run of MUSE and how it maintains the best accuracy (and the initial point) as it continues to search.

\subsubsection{Regression Results}
\begin{figure}[ht]
    \centering
    \includegraphics[width=\linewidth]{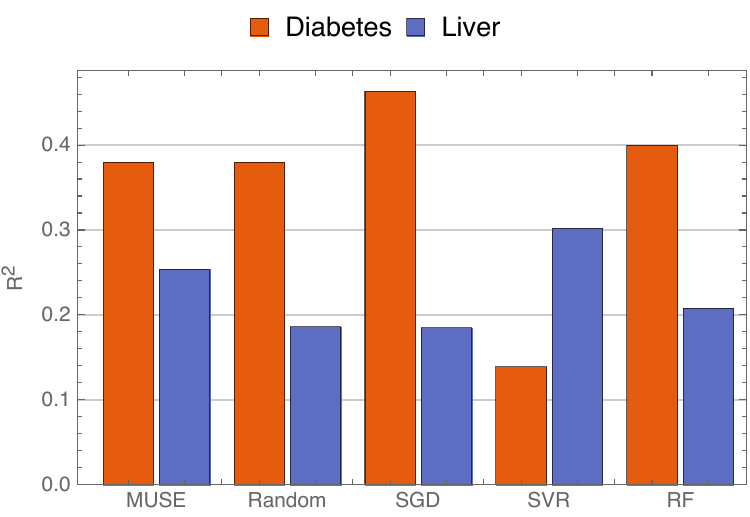}
    \caption{Comparison of the $R^2$ performance of the QVRs trained with MUSE, the QVRs trained with Random search and the classical regressors.}
    \label{fig:regres}
\end{figure}

Figure \ref{fig:regres} demonstrates that the quantum variational regressors (QVRs) trained with MUSE perform competitively and on par with the QVRs trained with Random search, and the classical regressors in terms of the coefficients of determination, that is, $R^2$ scores, with the two regression datasets. The types of regression models we evaluate are based on linear stochastic gradient descent (SGD), SVMs, and random forests (RFs) - the same types of the classification models.

\begin{table*}[ht]
    \centering
    \caption{Regression: Best and worst performing parameter combinations}
    \label{tab:Regression_bestworstparams}
    \begin{tabular}{|l|l|l|l|l|l|l|l|l|}  
     \hline
     &
     \multicolumn{4}{|c|}{\textbf{Best Parameters}} &
     \multicolumn{4}{|c|}{\textbf{Worst Parameters} } \\ \hline
        Dataset  &  ZZ Map Reps. & Ansatz Reps. & Scaler & Reducer &  
 ZZ Map Reps.  & Ansatz Reps. & Scaler & Reducer \\ \hline
       \textbf{Diabetes (Regression)}  & 1  & 3 &  MinMax & FA & 1  & 3 &  Standard & PCA  \\ \hline
       \textbf{Liver}  &  1  & 3 &  MinMax & FA & 1  & 2 &  MinMax & PCA   \\ \hline
    \end{tabular}    
\end{table*}

\begin{table}[ht]
    \centering
    \caption{Regression Scores ($R^2$) of MUSE and Random Search}
    \label{tab:minmaxR2scores}
    \begin{tabular}{|l|l|l|l|}  
     \hline
          &  MUSE  & Random Search  & Lowest Observed \\ \hline
       \textbf{Diabetes}  & 0.38  & 0.38  & -2.53  \\ \hline
       \textbf{Liver}  &  0.254  & 0.186 & -1.43  \\ \hline
    \end{tabular} 
\end{table}

Table \ref{tab:Regression_bestworstparams} details the best- and worst-performing parameter combinations in our regression evaluation. We see that similar to classification, the MinMax scaler performs better than the Standard scaler. We also see that FA-based feature reduction leads to higher $R^2$ scores than PCA.
We note that for the regression simulations, the numbers of the repetitions tested for a feature map are 1 and 2, and for an ansatz are 2 and 3 - different than the classification simulations.

Table \ref{tab:minmaxR2scores} shows the $R^2$ scores achieved by the QVRs trained with MUSE and the QVRs trained with Random search for the regression datasets. It also shows the observed lowest scores during the MUSE search. 
We see that MUSE improves the $R^2$ scores from being negative to positive for both datasets. 
When $R^2$ scores are negative, a model's prediction is arbitrarily incorrect. This is because that the assumed model is not a good fit to represent the data. Even always predicting the mean of the data has an $R^2$ score of zero. We conclude that MUSE significantly enhances the quality of the predictions of the QVRs. Lastly, we see that MUSE outperforms Random search with the Liver dataset, which is however limited.

\begin{figure}
    \centering
    \includegraphics[width=\linewidth]{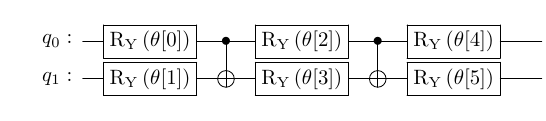}
    \caption{An Ansatz circuit in Qiskit \cite{Qiskit}.
    $q_0$ and $q_1$ are qubits. $R_Y$ are rotation gates about the $Y$-axis. $\theta$ is the parameters of the ansatz.}
    \label{fig:ansatz}
\end{figure}

To gain a more concrete understanding of quantum variational circuits, we include the following figures: Figure \ref{fig:zzmap} shows a feature map circuit and Figure \ref{fig:ansatz} shows an example ansatz circuit. 

\begin{figure*}
    \centering
    \includegraphics[width=\linewidth]{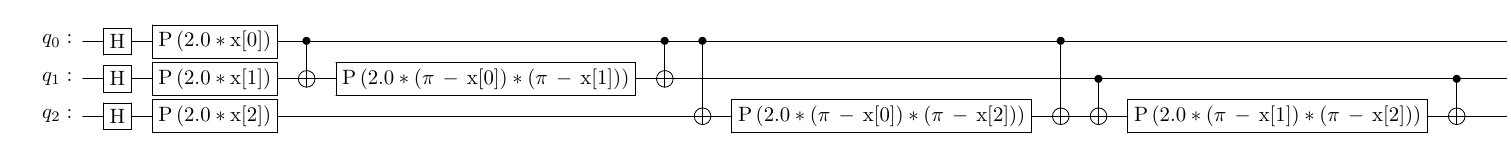}
    \caption{A ZZFeatureMap circuit with linear entanglement in Qiskit \cite{Qiskit}. $q_0$, $q_1$ and $q_2$ are qubits, Hadamart gates $H$ place
    qubits in superposition. $P$ gates are phase gates that rotate an input about the $Z$-axis. $x$ is the classical data features.}
    \label{fig:zzmap}
\end{figure*}

\subsection{Failed Attempt I: Classical Pretraining with Extreme \\ Machines}
In this section, we discuss one of our earlier attempts to find initial points that may lead to better performance for quantum variational learning. This idea was to use Extreme Machines \cite{huang2006extreme} to compute the initial classical points that boost the performance of quantum variational learners.  

Extreme machines \cite{huang2006extreme} \cite{wang2022review} are feed-forward neural networks that typical have one single hidden layer. Extreme machines do not perform gradient-based backpropagation. They use Moore-Penrose generalized inverse to compute and set the weights. They are single-iteration networks initialized with random inputs. These properties make them extremely fast, however, at the cost of being less accurate compared to widely used multilayer and deep neural networks with backpropagation. 

We implemented an extreme machine to compute the initial points for the quantum variational classifiers. However, many simulations showed that there was no consistent and sustained performance improvement. The simulations demonstrated that the accuracy of the quantum classifiers did not improve. As a result, we discontinued this idea.

\subsection{Failed Attempt II: Classical Ensembles of Quantum \\ Classifiers}
Other than extreme machines, we explored the idea of the classical ensembles of quantum variational classifiers. The idea was to randomly initialize several independent quantum classifiers and run them to get their predictions, i.e, classifications. After that, the idea was to use classical voting on the outputs of the quantum classifiers to classify each data instance. We performed many simulations with different datasets, where we used the majority-voted class as the finally predicted class. However, during the simulations, we did not observe any consistent improvement in detection accuracy. 
We aimed to use classical ensembles to improve the accuracy performance even though the training and inference costs would increase linearly with the number of independently trained quantum classifiers. In the end, this idea also failed to produce any performance improvement.

\section{Related Work}
\label{RelatedWork}
Quantum machine learning \cite{QuantumEnhancedMachine, schuld2018supervised, havlivcek2019supervised} consists of types of models that are conceptually the same as classical machine learning. Some examples are quantum kernel methods \cite{schuld2021supervised}, quantum neural networks \cite{abbas2021power}, quantum generative adversarial network \cite{huang2021experimental, stein2021qugan}, and quantum ensembles \cite{abbas2020quantum, stein2022eqc, macaluso2022quantum}.

In this study, we explore quantum variational learning \cite{benedetti2019parameterized, Variationallearningartificial, cerezo2021variational} which is a hybrid quantum and classical learning approach based on parameterised circuits. These parameterised circuits can be employed as quantum neural networks. For instance,
GenQu \cite{GenQu} is such a hybrid framework.
Similar to our work, it uses feature reduction based on PCA to reduce the number of the parameters of quantum neural networks. Compared to GenQu, our work focuses on automating quantum variational learning. We achieve this by offering MUSE as an initial point and parameter search algorithm for variational circuits. 
Another example of the usage of variational circuits for machine learning tasks is \cite{lockwood2020reinforcement}. The authors utilize these circuits and design a deep Q reinforcement learning framework.

While we design MUSE to perform structured searches for initial points and parameters that can help with the barren plateau problem \cite{mcclean2018barren, cerezo2021cost, marrero2021entanglement}, a focused in-depth (theoretical) analysis is out of scope of our work. To address this problem, 
there have been several recent works that offer tailored initialization techniques to  \cite{rad2022surviving, NEURIPS2022_7611a3cb, BEINIT, liu2021parameter, sauvage2021flip, grant2019initialization}. The authors of \cite{rad2022surviving} propose Bayesian optimization for parameter initialization. Zhang at. al. \cite{NEURIPS2022_7611a3cb} study Gaussian initialization both theoretically and empirically. They show that the gradient decreases polynomially (instead of exponentially) as the number of qubits and the circuit depth increase when Gaussian initialization is employed. Similarly, BEINIT \cite{BEINIT} is a initialization strategy based on perturbed Beta distribution. Compared to these initialization methods, we offer MUSE as a search strategy visiting multiple neighborhoods of the value space in a structured way. Additionally, as it searches for the best-performing initial points, MUSE evaluates the circuit parameters and data preprocessing options at the same time. Other studies such as \cite{sauvage2021flip, grant2019initialization, liu2021parameter} are based on additional classical learners for the initial parameters or transfer learning. Differently, MUSE searches the value space of the initial parameters.

In quantum variational learning, architecture (circuit) search \cite{du2022quantum, du2020quantum, NEURIPS2021_97244127, gomez2022towards, QuantumNAS, kuo2021quantum, zhang2022evolutionary, EvolutionaryQuantum, duong2022quantum, rad2022surviving} is recently getting much attention. The variational circuit search approaches cover a wide range of methods.
These methods can be based on evolutionary algorithms \cite{zhang2022evolutionary, EvolutionaryQuantum}, reinforcement learning \cite{NEURIPS2021_97244127, kuo2021quantum}, Bayesian optimization \cite{duong2022quantum}. Comparing to our work, variational circuit search is closely related. While quantum architecture search primarily seeks the best-performing quantum circuits, we seek the best-performing initial points, circuit parameters, and classical data preprocessing options. The parameters we explore do define a variational circuit which is how our work overlaps with circuit search. However, our primary focus is not the variational circuit, it is to find the best-performing initial point to a \emph{selected} variational circuit and classical data preprocessing options.

As we stated in Section \ref{Results}, we explored the idea of classical ensembles of quantum classifiers which did not lead to a success. 
There are studies that experiment with classical ensembles \cite{qin2022improving, silver2022quilt}. Qin et. al. \cite{qin2022improving} report that their classical ensemble which is based majority voting improves its accuracy with the MNIST dataset. In comparison, we did not observe any consistent accuracy improvement with our datasets.
QUILT \cite{silver2022quilt} is another classical ensemble of quantum classifiers. It determines the class of an instance based on the weights of the individual quantum classifiers. These weights are calculated simply by the ratio of a quantum classifier's accuracy over the sum of all accuracies of the classifiers multiplied by the ensemble confidence score. The authors report that QUILT achieved improved accuracy over the individual classifiers. In the future, we plan to evaluate their confidence score based approach on our datasets. 
In addition to classical ensembles, quantum ensembles of quantum classifiers have also been studied \cite{schuld2018quantumens, stein2022eqc, macaluso2022quantum, abbas2020quantum}. For instance, Stein et. al. \cite{stein2022eqc} develop a distributed framework that performs parallel training of variational quantum algorithms. Another example is that Macaluso et. al. \cite{macaluso2022quantum} introduce a classification algorithm based on quantum ensembles built by bagging. These proposed ensembles are not classical. That is, the ensemble itself is either a fully quantum or a hybrid quantum algorithm. 

\section{Conclusions}
\label{Conclusion}
In this work, we devise and develop a parallelizable multi-locality search algorithm, named MUSE, for quantum variational learning to find the initial points and the combinations of parameters and data preprocessing options to achieve the best possible performance. 
MUSE searches multiple localities in the (multi-dimensional) space of the initial points to quantum variational circuits. Simulations with a total of seven real-world datasets demonstrate that MUSE significantly improves the performance of quantum variational learning with respect to the observed lowest scores in both classification and regression tasks. In classification, we see that the average accuracy improvement is 2.3$\times$ with respect to the observed lowest accuracy scores during a search. In addition, the detection performance is on par with classical classification.
Furthermore, in regression, we see that the performance in terms of $R^2$ scores significantly increases with MUSE, and improves from negative scores to positive scores. MUSE achieves similar $R^2$ scores with classical regression.

In the future, we plan to use the IBM quantum devices to perform simulations with MUSE on real hardware. Due to lack of funding, we could not perform such simulations.
We also aim to explore other aspects and parameters such as different feature maps and optimizers already available in Qiskit.

\bibliographystyle{unsrt} 
\bibliography{refs.bib} 

\begin{thebibliography}{10}

\bibitem{rieffel2011quantum}
Eleanor~G Rieffel and Wolfgang~H Polak.
\newblock {\em Quantum computing: A gentle introduction}.
\newblock MIT Press, 2011.

\bibitem{harrow2017quantum}
Aram~W Harrow and Ashley Montanaro.
\newblock Quantum computational supremacy.
\newblock {\em Nature}, 549(7671):203--209, 2017.

\bibitem{dong2021survey}
Shi Dong, Ping Wang, and Khushnood Abbas.
\newblock A survey on deep learning and its applications.
\newblock {\em Computer Science Review}, 40:100379, 2021.

\bibitem{zhang2020recent}
Yao Zhang and Qiang Ni.
\newblock Recent advances in quantum machine learning.
\newblock {\em Quantum Engineering}, 2(1):e34, 2020.

\bibitem{cerezo2022challenges}
M~Cerezo, Guillaume Verdon, Hsin-Yuan Huang, Lukasz Cincio, and Patrick~J
  Coles.
\newblock Challenges and opportunities in quantum machine learning.
\newblock {\em Nature Computational Science}, 2(9):567--576, 2022.

\bibitem{cerezo2021variational}
Marco Cerezo, Andrew Arrasmith, Ryan Babbush, Simon~C Benjamin, Suguru Endo,
  Keisuke Fujii, Jarrod~R McClean, Kosuke Mitarai, Xiao Yuan, Lukasz Cincio,
  et~al.
\newblock Variational quantum algorithms.
\newblock {\em Nature Reviews Physics}, 3(9):625--644, 2021.

\bibitem{gyongyosi2019survey}
Laszlo Gyongyosi and Sandor Imre.
\newblock A survey on quantum computing technology.
\newblock {\em Computer Science Review}, 31:51--71, 2019.

\bibitem{du2022quantum}
Yuxuan Du, Tao Huang, Shan You, Min-Hsiu Hsieh, and Dacheng Tao.
\newblock Quantum circuit architecture search for variational quantum
  algorithms.
\newblock {\em npj Quantum Information}, 8(1):62, 2022.

\bibitem{du2020quantum}
Yuxuan Du, Tao Huang, Shan You, Min-Hsiu Hsieh, and Dacheng Tao.
\newblock Quantum circuit architecture search: error mitigation and
  trainability enhancement for variational quantum solvers.
\newblock {\em arXiv preprint arXiv:2010.10217}, 2020.

\bibitem{NEURIPS2021_97244127}
Mateusz Ostaszewski, Lea~M. Trenkwalder, Wojciech Masarczyk, Eleanor Scerri,
  and Vedran Dunjko.
\newblock Reinforcement learning for optimization of variational quantum
  circuit architectures.
\newblock In M.~Ranzato, A.~Beygelzimer, Y.~Dauphin, P.S. Liang, and J.~Wortman
  Vaughan, editors, {\em Advances in Neural Information Processing Systems},
  volume~34, pages 18182--18194. Curran Associates, Inc., 2021.

\bibitem{gomez2022towards}
Ra{\'u}l~Berganza G{\'o}mez, Corey O’Meara, Giorgio Cortiana, Christian~B
  Mendl, and Juan Bernab{\'e}-Moreno.
\newblock Towards autoqml: A cloud-based automated circuit architecture search
  framework.
\newblock In {\em 2022 IEEE 19th International Conference on Software
  Architecture Companion (ICSA-C)}, pages 129--136. IEEE, 2022.

\bibitem{QuantumNAS}
Hanrui Wang, Yongshan Ding, Jiaqi Gu, Yujun Lin, David~Z. Pan, Frederic~T.
  Chong, and Song Han.
\newblock Quantumnas: Noise-adaptive search for robust quantum circuits.
\newblock In {\em 2022 IEEE International Symposium on High-Performance
  Computer Architecture (HPCA)}, pages 692--708, 2022.

\bibitem{kuo2021quantum}
En-Jui Kuo, Yao-Lung~L Fang, and Samuel Yen-Chi Chen.
\newblock Quantum architecture search via deep reinforcement learning.
\newblock {\em arXiv preprint arXiv:2104.07715}, 2021.

\bibitem{zhang2022evolutionary}
Anqi Zhang and Shengmei Zhao.
\newblock Evolutionary-based quantum architecture search.
\newblock {\em arXiv preprint arXiv:2212.00421}, 2022.

\bibitem{EvolutionaryQuantum}
Li~Ding and Lee Spector.
\newblock Evolutionary quantum architecture search for parametrized quantum
  circuits.
\newblock In {\em Proceedings of the Genetic and Evolutionary Computation
  Conference Companion}, GECCO '22, page 2190–2195, New York, NY, USA, 2022.
  Association for Computing Machinery.

\bibitem{duong2022quantum}
Trong Duong, Sang~T Truong, Minh Tam, Bao Bach, Ju-Young Ryu, and
  June-Koo~Kevin Rhee.
\newblock Quantum neural architecture search with quantum circuits metric and
  bayesian optimization.
\newblock {\em arXiv preprint arXiv:2206.14115}, 2022.

\bibitem{schuld2021machine}
Maria Schuld and Francesco Petruccione.
\newblock {\em Machine learning with quantum computers}.
\newblock Springer, 2021.

\bibitem{schuld2018supervised}
Maria Schuld and Francesco Petruccione.
\newblock {\em Supervised learning with quantum computers}, volume~17.
\newblock Springer, 2018.

\bibitem{sim2019expressibility}
Sukin Sim, Peter~D Johnson, and Al{\'a}n Aspuru-Guzik.
\newblock Expressibility and entangling capability of parameterized quantum
  circuits for hybrid quantum-classical algorithms.
\newblock {\em Advanced Quantum Technologies}, 2(12):1900070, 2019.

\bibitem{Qiskit}
{Qiskit contributors}.
\newblock Qiskit: An open-source framework for quantum computing, 2023.

\bibitem{Dua2019}
Dheeru Dua and Casey Graff.
\newblock {UCI} machine learning repository.
\newblock \url{http://archive.ics.uci.edu/ml}, 2017.

\bibitem{yeh2009knowledge}
I-Cheng Yeh, King-Jang Yang, and Tao-Ming Ting.
\newblock Knowledge discovery on rfm model using bernoulli sequence.
\newblock {\em Expert Systems with Applications}, 36(3):5866--5871, 2009.

\bibitem{mcdermott2016diagnosing}
James McDermott and Richard~S Forsyth.
\newblock Diagnosing a disorder in a classification benchmark.
\newblock {\em Pattern Recognition Letters}, 73:41--43, 2016.

\bibitem{Efron2004}
Bradley Efron, Trevor Hastie, Iain Johnstone, and Robert Tibshirani.
\newblock Least angle regression.
\newblock {\em The Annals of Statistics}, 32(2), 2004.

\bibitem{ScikitLearnDiabetes}
Scikit-Learn.
\newblock Scikit-learn diabetes dataset.
\newblock
  \url{https://scikit-learn.org/stable/modules/generated/sklearn.datasets.load\_diabetes.html},
  Accessed on 04-06-2023.

\bibitem{huang2006extreme}
Guang-Bin Huang, Qin-Yu Zhu, and Chee-Kheong Siew.
\newblock Extreme learning machine: theory and applications.
\newblock {\em Neurocomputing}, 70(1-3):489--501, 2006.

\bibitem{wang2022review}
Jian Wang, Siyuan Lu, Shui-Hua Wang, and Yu-Dong Zhang.
\newblock A review on extreme learning machine.
\newblock {\em Multimedia Tools and Applications}, 81(29):41611--41660, 2022.

\bibitem{QuantumEnhancedMachine}
Vedran Dunjko, Jacob~M. Taylor, and Hans~J. Briegel.
\newblock Quantum-enhanced machine learning.
\newblock {\em Phys. Rev. Lett.}, 117:130501, Sep 2016.

\bibitem{havlivcek2019supervised}
Vojt{\v{e}}ch Havl{\'\i}{\v{c}}ek, Antonio~D C{\'o}rcoles, Kristan Temme,
  Aram~W Harrow, Abhinav Kandala, Jerry~M Chow, and Jay~M Gambetta.
\newblock Supervised learning with quantum-enhanced feature spaces.
\newblock {\em Nature}, 567(7747):209--212, 2019.

\bibitem{schuld2021supervised}
Maria Schuld.
\newblock Supervised quantum machine learning models are kernel methods.
\newblock {\em arXiv preprint arXiv:2101.11020}, 2021.

\bibitem{abbas2021power}
Amira Abbas, David Sutter, Christa Zoufal, Aur{\'e}lien Lucchi, Alessio
  Figalli, and Stefan Woerner.
\newblock The power of quantum neural networks.
\newblock {\em Nature Computational Science}, 1(6):403--409, 2021.

\bibitem{huang2021experimental}
He-Liang Huang, Yuxuan Du, Ming Gong, Youwei Zhao, Yulin Wu, Chaoyue Wang,
  Shaowei Li, Futian Liang, Jin Lin, Yu~Xu, et~al.
\newblock Experimental quantum generative adversarial networks for image
  generation.
\newblock {\em Physical Review Applied}, 16(2):024051, 2021.

\bibitem{stein2021qugan}
Samuel~A Stein, Betis Baheri, Daniel Chen, Ying Mao, Qiang Guan, Ang Li,
  Bo~Fang, and Shuai Xu.
\newblock Qugan: A quantum state fidelity based generative adversarial network.
\newblock In {\em 2021 IEEE International Conference on Quantum Computing and
  Engineering (QCE)}, pages 71--81. IEEE, 2021.

\bibitem{abbas2020quantum}
Amira Abbas, Maria Schuld, and Francesco Petruccione.
\newblock On quantum ensembles of quantum classifiers.
\newblock {\em Quantum Machine Intelligence}, 2:1--8, 2020.

\bibitem{stein2022eqc}
Samuel Stein, Nathan Wiebe, Yufei Ding, Peng Bo, Karol Kowalski, Nathan Baker,
  James Ang, and Ang Li.
\newblock Eqc: ensembled quantum computing for variational quantum algorithms.
\newblock In {\em Proceedings of the 49th Annual International Symposium on
  Computer Architecture}, pages 59--71, 2022.

\bibitem{macaluso2022quantum}
Antonio Macaluso, Luca Clissa, Stefano Lodi, and Claudio Sartori.
\newblock Quantum ensemble for classification, 2022.

\bibitem{benedetti2019parameterized}
Marcello Benedetti, Erika Lloyd, Stefan Sack, and Mattia Fiorentini.
\newblock Parameterized quantum circuits as machine learning models.
\newblock {\em Quantum Science and Technology}, 4(4):043001, 2019.

\bibitem{Variationallearningartificial}
Francesco Tacchino, Panagiotis~Kl. Barkoutsos, Chiara Macchiavello, Dario
  Gerace, Ivano Tavernelli, and Daniele Bajoni.
\newblock Variational learning for quantum artificial neural networks.
\newblock In {\em 2020 IEEE International Conference on Quantum Computing and
  Engineering (QCE)}, pages 130--136, 2020.

\bibitem{GenQu}
Samuel~A Stein, Ryan L’Abbate, Wenrui Mu, Yue Liu, Betis Baheri, Ying Mao,
  Guan Qiang, Ang Li, and Bo~Fang.
\newblock A hybrid system for learning classical data in quantum states.
\newblock In {\em 2021 IEEE International Performance, Computing, and
  Communications Conference (IPCCC)}, pages 1--7, 2021.

\bibitem{lockwood2020reinforcement}
Owen Lockwood and Mei Si.
\newblock Reinforcement learning with quantum variational circuit.
\newblock In {\em Proceedings of the AAAI Conference on Artificial Intelligence
  and Interactive Digital Entertainment}, volume~16, pages 245--251, 2020.

\bibitem{mcclean2018barren}
Jarrod~R McClean, Sergio Boixo, Vadim~N Smelyanskiy, Ryan Babbush, and Hartmut
  Neven.
\newblock Barren plateaus in quantum neural network training landscapes.
\newblock {\em Nature communications}, 9(1):4812, 2018.

\bibitem{cerezo2021cost}
Marco Cerezo, Akira Sone, Tyler Volkoff, Lukasz Cincio, and Patrick~J Coles.
\newblock Cost function dependent barren plateaus in shallow parametrized
  quantum circuits.
\newblock {\em Nature communications}, 12(1):1791, 2021.

\bibitem{marrero2021entanglement}
Carlos~Ortiz Marrero, M{\'a}ria Kieferov{\'a}, and Nathan Wiebe.
\newblock Entanglement-induced barren plateaus.
\newblock {\em PRX Quantum}, 2(4):040316, 2021.

\bibitem{rad2022surviving}
Ali Rad, Alireza Seif, and Norbert~M Linke.
\newblock Surviving the barren plateau in variational quantum circuits with
  bayesian learning initialization.
\newblock {\em arXiv preprint arXiv:2203.02464}, 2022.

\bibitem{NEURIPS2022_7611a3cb}
Kaining Zhang, Liu Liu, Min-Hsiu Hsieh, and Dacheng Tao.
\newblock Escaping from the barren plateau via gaussian initializations in deep
  variational quantum circuits.
\newblock In S.~Koyejo, S.~Mohamed, A.~Agarwal, D.~Belgrave, K.~Cho, and A.~Oh,
  editors, {\em Advances in Neural Information Processing Systems}, volume~35,
  pages 18612--18627. Curran Associates, Inc., 2022.

\bibitem{BEINIT}
Ankit Kulshrestha and Ilya Safro.
\newblock Beinit: Avoiding barren plateaus in variational quantum algorithms.
\newblock In {\em 2022 IEEE International Conference on Quantum Computing and
  Engineering (QCE)}, pages 197--203, 2022.

\bibitem{liu2021parameter}
Huan-Yu Liu, Tai-Ping Sun, Yu-Chun Wu, Yong-Jian Han, and Guo-Ping Guo.
\newblock A parameter initialization method for variational quantum algorithms
  to mitigate barren plateaus based on transfer learning.
\newblock {\em arXiv preprint arXiv:2112.10952}, 2021.

\bibitem{sauvage2021flip}
Frederic Sauvage, Sukin Sim, Alexander~A Kunitsa, William~A Simon, Marta Mauri,
  and Alejandro Perdomo-Ortiz.
\newblock Flip: A flexible initializer for arbitrarily-sized parametrized
  quantum circuits.
\newblock {\em arXiv preprint arXiv:2103.08572}, 2021.

\bibitem{grant2019initialization}
Edward Grant, Leonard Wossnig, Mateusz Ostaszewski, and Marcello Benedetti.
\newblock An initialization strategy for addressing barren plateaus in
  parametrized quantum circuits.
\newblock {\em Quantum}, 3:214, 2019.

\bibitem{qin2022improving}
Ruiyang Qin, Zhiding Liang, Jinglei Cheng, Peter Kogge, and Yiyu Shi.
\newblock Improving quantum classifier performance in nisq computers by voting
  strategy from ensemble learning, 2022.

\bibitem{silver2022quilt}
Daniel Silver, Tirthak Patel, and Devesh Tiwari.
\newblock Quilt: Effective multi-class classification on quantum computers
  using an ensemble of diverse quantum classifiers.
\newblock In {\em Proceedings of the AAAI Conference on Artificial
  Intelligence}, volume~36, pages 8324--8332, 2022.

\bibitem{schuld2018quantumens}
Maria Schuld and Francesco Petruccione.
\newblock Quantum ensembles of quantum classifiers.
\newblock {\em Scientific reports}, 8(1):2772, 2018.

\end{thebibliography}

\end{document}